\documentclass[10pt,twocolumn,letterpaper]{article}

\usepackage[pagenumbers]{cvpr} 

\usepackage[dvipsnames]{xcolor}
\usepackage[utf8]{inputenc} 
\usepackage[T1]{fontenc}    
\usepackage{booktabs}       
\usepackage{amsfonts}       
\usepackage{amsmath}
\usepackage{amsthm}
\usepackage{nicefrac}       
\usepackage{microtype}      
\usepackage{graphicx}
\usepackage{graphbox}
\usepackage{subcaption}
\usepackage{algorithm}
\usepackage{algpseudocode}
\usepackage{bm}
\usepackage{colortbl,array}
\usepackage{multirow}
\usepackage{makecell}
\usepackage{listings}

\usepackage{pifont}
\usepackage{rotating}

\newcommand{\TODO}[1]{}
\renewcommand{\TODO}[1]{{\color{cyan} [TODO: {#1}]}}
\newcommand{\WIP}[1]{}
\renewcommand{\WIP}[1]{{\color{magenta} [WIP: {#1}]}}
\definecolor{midblue}{rgb}{0,0.11372549,0.258823529}
\newcommand{\argmax}{\mathop{\rm arg~max}\limits}

\definecolor{codegray}{rgb}{0.5,0.5,0.5}
\definecolor{keycolor}{rgb}{0.13, 0.29, 0.53} 
\definecolor{stringcolor}{rgb}{0.31, 0.60, 0.02} 
\definecolor{backcolour}{rgb}{0.98,0.98,0.98}

\lstdefinelanguage{json}{
    basicstyle=\small\ttfamily,
    numbers=left,
    numberstyle=\tiny\color{codegray},
    stepnumber=1,
    numbersep=5pt,
    showstringspaces=false,
    breaklines=true,
    frame=lines,
    backgroundcolor=\color{backcolour},
    literate=
     *{0}{{{\color{orange}0}}}{1}
      {1}{{{\color{orange}1}}}{1}
      {2}{{{\color{orange}2}}}{1}
      {3}{{{\color{orange}3}}}{1}
      {4}{{{\color{orange}4}}}{1}
      {5}{{{\color{orange}5}}}{1}
      {6}{{{\color{orange}6}}}{1}
      {7}{{{\color{orange}7}}}{1}
      {8}{{{\color{orange}8}}}{1}
      {9}{{{\color{orange}9}}}{1}
      {:}{{{\color{red}{:}}}}{1}
      {,}{{{\color{red}{,}}}}{1}
      {\{}{{{\color{black}{\{}}}}{1}
      {\}}{{{\color{black}{\}}}}}{1}
      {[}{{{\color{black}{[}}}}{1}
      {]}{{{\color{black}{]}}}}{1},
    morestring=[b]",
    stringstyle=\color{stringcolor},
    keywordstyle=\color{keycolor}\bfseries,
}

\usepackage[most]{tcolorbox}
\tcbuselibrary{listings}
\definecolor{lightgray}{gray}{0.95}

\newtcolorbox{promptbox}[1][]{
  colback=lightgray, 
  colframe=gray,     
  boxrule=0.5pt,     
  arc=1mm,           
  fontupper=\scriptsize\ttfamily, 
  title={Prompt},    
  #1
}

\definecolor{LightRed}{rgb}{0.99,0.91,0.91}    
\definecolor{LightBlue}{rgb}{0.88,0.92,0.95}   
\definecolor{LightYellow}{rgb}{0.98,0.96,0.89} 

\definecolor{cvprblue}{rgb}{0.21,0.49,0.74}
\usepackage[pagebackref,breaklinks,colorlinks,allcolors=cvprblue]{hyperref}
\newcommand{\ptab}{\hspace*{1em}}

\def\confName{CVPR}
\def\confYear{2027}

\title{Relational Bottleneck Models}

\author{Shinya' Yamaguchi\\
NTT\\
{\tt\small shinya.yamaguchi@ntt.com}
\and
Kosuke Nishida\\
NTT\\
{\tt\small kosuke.nishida@ntt.com}
}

\begin{document}
\maketitle

\begin{abstract}
Achieving both interpretable and high-performance inference is a key challenge in deep learning.
While Concept Bottleneck Models (CBMs) provide inherent interpretability, they are constrained by task-specific training and rely on isolated visual concepts that fail to capture higher-order visual semantic relationships.
Conversely, Large Vision Language Models (LVLMs) demonstrate zero-shot capabilities through Chain-of-Thought (CoT) reasoning.
However, standard CoT lacks a strict causal dependency between the generated rationale and the final output, compromising faithfulness.
To bridge this gap, we propose Relational Bottleneck Models (RBMs), a novel training-free framework that transforms any off-the-shelf LVLMs into inherently interpretable models in a plug-and-play manner.
RBMs consist of two stages: (i) Structured Rationale Generation (SRG), which explicitly models semantic interactions between visual concepts as subject-predicate-object triplets, and (ii) Rationale-Grounded Decoding (RGD), which enforces a strict causal dependency between the rationales and the final output.
RGD aggregates rationale-conditional next-token distributions via an importance-weighted product-of-experts strategy, strictly excluding the image inputs from the decoding context.
Extensive zero-shot evaluations across diverse benchmarks and LVLM backbones demonstrate that RBMs achieve superior interpretability while improving task performance over CBM baselines, entirely without task-specific fine-tuning.\looseness-1
\end{abstract}

\begin{figure}[tbp]
    \centering
    \includegraphics[width=\linewidth]{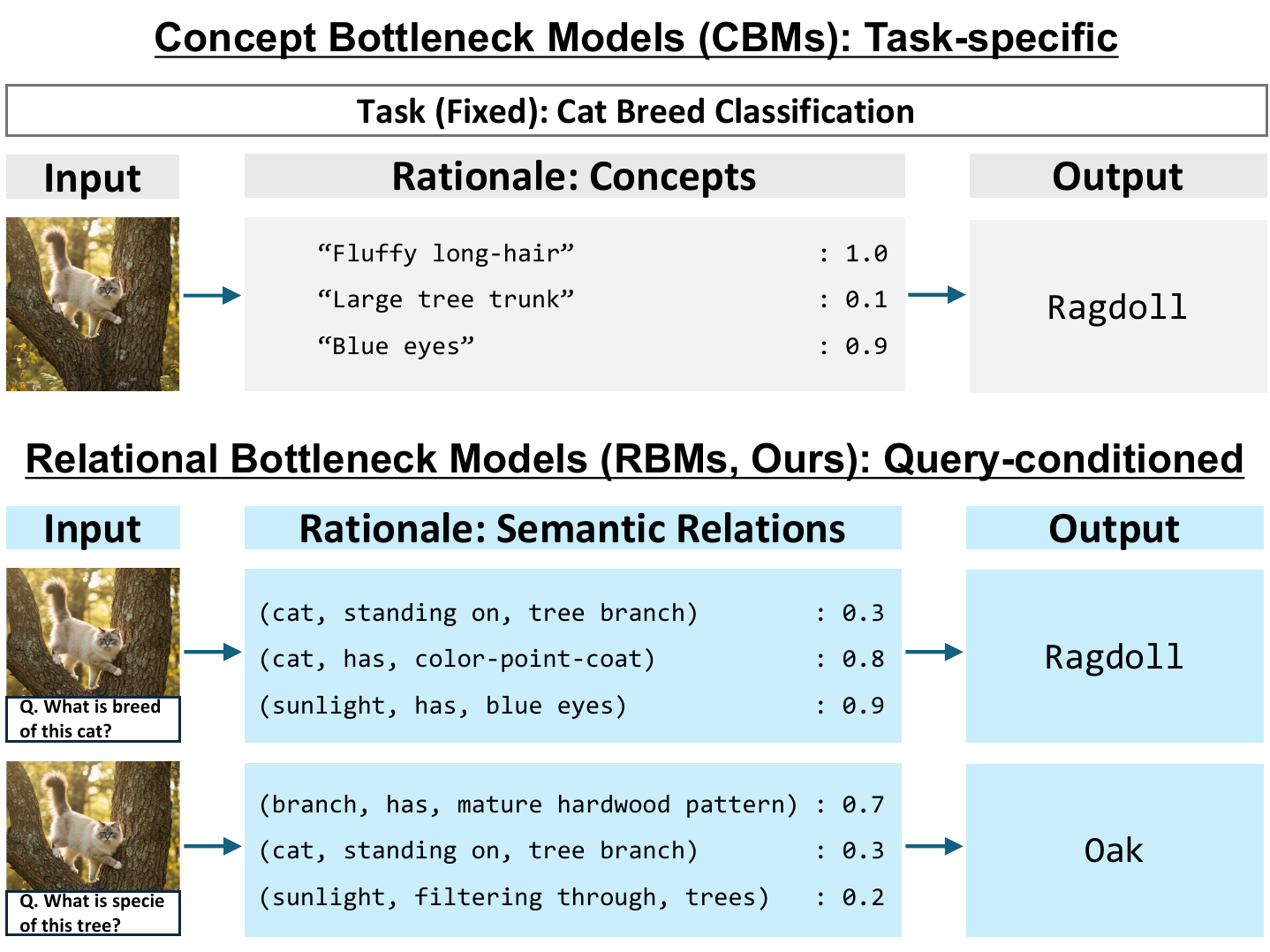}
    \caption{
    \textbf{Overview.} We propose an interpretable model, Relational Bottleneck Models (RBMs), based on large vision language models (LVLMs) in a plug-and-play manner.
    Previous Concept Bottleneck Models (CBMs) offer limited interpretability through their isolated concepts and are often limited in their applicability to the fixed task during training.
    In contrast, our RBMs can provide informative rationales, formed by visual semantic relation triplets with self-verbalized importance scores, and can be flexibly applied to various tasks that backbone LVLMs can solve without additional training.
    }
    \label{fig:top}
    \vspace{-5mm}
\end{figure}

\section{Introduction}\label{sec:intro}
Achieving interpretable inference with deep learning models remains a major research interest~\cite{Li_2022_interpretable-dl-survey,  Singh_2024_interpretability-llm-survey, Dang_2024_interpretability-mllm-survey}.
Since deep models typically perform as black boxes, understanding the decision-making process behind their final predictions is challenging.
To address this, existing literature has attempted to build \textit{inherently interpretable models}, which explicitly generate the rationale for a decision as an intermediate output~\cite{Singh_2024_interpretability-llm-survey, Dang_2024_interpretability-mllm-survey}.
A representative class of inherently interpretable models is Concept Bottleneck Models (CBMs,~\citet{Koh_ICML20_concept_bottleneck}).
CBMs first generate rationales consisting of ``concepts'', representing visual components of objects or classes, from an input image, and subsequently predict the final task label based solely on these concepts.
By enforcing that the prediction causally depends on this concept bottleneck, CBMs ensure the interpretability of their outputs.
Furthermore, CBMs enable human intervention, allowing users to debug the model by correcting mispredicted concepts to steer the final output~\cite{Koh_ICML20_concept_bottleneck}.
However, CBMs face significant limitations in terms of expressiveness and generalizability.
First, they typically rely on the independent contributions of pre-defined low-level concepts, making it difficult to capture complex semantic relationships among concepts~\cite{Raman_ICML24_understanding-inter-concept}.
Second, CBMs are often designed for specific tasks (e.g., classification) and require task-specific training, hindering their transferability across diverse tasks.

Recently, Large Vision-Language Models (LVLMs) have emerged as a powerful paradigm~\cite{liu_NeurIPS23_llava, Google_2025_gemma3, Wang_2025_internvl3.5, Bai_2025_qwen3-vl}. 
LVLMs achieve remarkable zero-shot inference capability across various tasks and enable user interaction via visual instruction tuning.
Notably, LVLMs can generate reasoning paths or structured rationale scene graphs via multi-modal Chain-of-Thought (CoT), which has been shown to improve inference accuracy~\cite{Zhang_TMLR_mm-cot, Mitra_CVPR24_CCoT, Tang_CVPR25_reason-before-retrieve-mm-cot}.
Despite these advancements, CoT suffers from fatal shortcomings in terms of interpretability~\cite{Jie_NAACL24_interpretability_cot, Barez_2025_cot_is_not_explainability, Yamaguchi_2025_red-mm-cot} and faithfulness~\cite{Turpin_NeurIPS23_cot-faithfulness, Paul_EMNLP24_improving-cot-faithfulness}.
Since multi-modal CoT in LVLMs is implemented via autoregressive next-token prediction conditioned on an image, a question, and the generated rationales, it lacks a strict causal dependency between rationales and final outputs, unlike in CBMs.
In other words, there is no structural guarantee that the final output is causally generated from the produced rationale; the model may simply ignore the rationale and overly rely on the image features within its input context.
Indeed, prior work~\cite{Yamaguchi_2025_red-mm-cot} has observed that LVLMs often prioritize visual tokens over CoT rationales when forming final predictions. Consequently, while LVLMs achieve superior zero-shot performance, they fail to provide the faithful interpretability and intervenability characteristic of CBMs, as the final output is not strictly grounded in the rationale.
In this paper, we aim to answer the following research question: \textit{Can LVLMs perform plug-and-play interpretable inference for arbitrary tasks?}\looseness-1

To address this question, we introduce \textbf{Relational Bottleneck Models (RBMs)}, a novel framework built on top of LVLMs that transforms any off-the-shelf LVLM into an inherently interpretable model in a plug-and-play manner.
Our framework consists of two stages: 
(i) \textbf{Structured Rationale Generation (SRG)} and (ii) \textbf{Rationale-Grounded Decoding (RGD)}.
SRG is formalized as producing a structured intermediate rationale that represents semantic relationships between visual concepts in the image as \textit{subject-predicate-object} triplets inferred by the multi-modal CoT.
Compared to the independent concepts in CBMs, SRG captures higher-order interactions that directly represent the cause of the final output, providing more meaningful interpretability (Fig.~\ref{fig:top}).
Furthermore, we find that SRG achieves higher inference performance compared to standard concepts and unstructured reasoning text.
To ensure the final prediction is grounded in the generated rationale, RGD enforces the use of the rationales during the next-token prediction for the final output.
Specifically, RGD constructs a prediction distribution conditioned on each extracted semantic relation triplet and aggregates them into an ensemble distribution using a product-of-experts strategy, weighted by self-verbalized importance scores from the LVLM.
Crucially, this ensemble distribution determines the final output based solely on the rationales, strictly excluding the image from the final decoding context.
This mechanism enhances a causal dependency between the rationale and the output, thereby ensuring interpretability.

We evaluate RBMs on multiple benchmarks, including classification, VQA, and captioning, using various LVLM backbones in a fully zero-shot manner.
Our experiments demonstrate that RBMs achieve superior interpretability while improving zero-shot performance over CBM baselines.
We believe that this work presents a new direction for constructing inherently interpretable models using arbitrary LVLMs, paving the way for the next generation of trustworthy foundation models.

\section{Related Work}\label{sec:related_work}
\noindent\textbf{Concept Bottleneck Models.}
Concept bottleneck models (CBMs)~\cite{Koh_ICML20_concept_bottleneck} are a representative family of inherently interpretable models that causally base their predictions on intermediate rationales called ``concepts'', which are visual components of objects or class categories within input images.
CBMs guarantee interpretability through a bi-level prediction scheme: image-to-concept and concept-to-class.
Early works on CBMs~\cite{Koh_ICML20_concept_bottleneck, Zarlenga_NeurIPS22_concept_embedding} directly learn the bi-level predictions via supervised objectives on a fully annotated dataset $\mathcal{D}$ with tuples of input image $x$, concepts $C=\{c_k\in\mathcal{C}\}_{k=1}^{K}$ and class label $y$:
\begin{equation}\label{eq:cbm_obj}
    \min_{g, h} \mathbb{E}_{(x,C,y)\in\mathcal{D}}\left[ \mathcal{L}(h\circ g(x),y) + \alpha\sum_{k}^K\mathcal{L}(g(x)_k,c_k)\right],
\end{equation}
where $g$ and $h$ denote the concept and class label predictors, and \(\mathcal{L}\) is a supervised loss function such as softmax cross-entropy loss.
While these original CBMs theoretically provide a rigorous mapping based on human annotations, they impose a massive annotation cost and often suffer from degraded target task performance due to the training difficulty on long-tail concept distributions~\cite{Ramaswamy_CVPR23_overlooked_factors_CBMs}.
Post-hoc CBMs~\cite{Yuksekgonul_ICLR23_post-hoc_CBMs}, Label-free CBMs~\cite{Oikarinen_ICLR23_label-free_CBMs}, and LaBo~\cite{Yang_CVPR23_labo} concurrently addressed these challenges by leveraging multi-modal encoders like CLIP~\cite{Radford_ICML21_CLIP} and Large Language Models (LLMs) to automatically construct a concept bank without manual annotations.
Building upon this breakthrough, subsequent works on CBMs have rapidly expanded.
Notable extensions include open-set classification~\cite{Tan_ECCV24_OpenCBM, Bhalla_NeurIPS24_splice, Yamaguchi_2025_zcbm, Zhang_CVPR25_attribute-formed-cbm, Zeng_CVPR25_domain-generalization-cbm}, dynamic concept identification~\cite{Chiquier_ECCV24_evolving_interpretable_visual_classifier, Shang_CVPR24_incremental_residual_CBMs, Liu_CVPR25_hybrid-cbm}, integration of logical expressions~\cite{Barbiero_ICML23_neural-symbolic-cbm}, fine-grained mapping to local visual features~\cite{Xie_CVPR25_discovering_fine-grained}, and adaptation to language models beyond vision models~\cite{Sun_ICLR25_CBLLM}.
However, in principle, these CBMs assign scalar scores for every single concept and do not consider the relationship among concepts, resulting in limited interpretability~\cite{Raman_ICML24_understanding-inter-concept, Xu_ICLR24_energy-based_CBMs, Yamaguchi_AAAI25_xbm}.
Furthermore, since CBMs are often designed for specific target tasks, e.g., classification, they cannot be generalized to arbitrary tasks in a zero-shot manner.
In contrast, our RBMs provide visual semantic relationships between concepts as rationales and powerful zero-shot capabilities across various tasks via a plug-and-play decoding algorithm on pre-trained LVLMs.

\noindent\textbf{Large Vision Language Models.}
Large Vision Language Models (LVLMs) are autoregressive text generative models composed of visual encoders and LLMs.
LVLMs achieve powerful zero-shot capabilities across diverse multi-modal tasks (e.g., VQA~\cite{Hudson_CVPR19_gqa}) by aligning visual embeddings with LLM token representations via visual instruction tuning~\cite{liu_NeurIPS23_llava, Liu_CVPR24_improved_llava_llavabench, Li_2024_llavanext, Li_2024_llava-ov, Google_2025_gemma3, Wang_2025_internvl3.5, Bai_2025_qwen3-vl}.
Similar to LLMs~\cite{Wei_NeurIPS22_CoT, Kojima_NeurIPS22_lets_think_step_by_step}, LVLMs can enhance their multi-modal reasoning capability by chain-of-thought (CoT) prompting~\cite{Zhang_TMLR_mm-cot, He_AAAI24_mm_latent_cot, Mondal_AAAI24_kam-cot, Zheng_LREC24_enhancing_mmcot_soft_negative_sampling, Mitra_CVPR24_CCoT, Tang_CVPR25_reason-before-retrieve-mm-cot, Xia_ICCV25_bootstrapping-grounded-mm-cot, Jiang_ICCV25_corvid-mm-cot, Sun_NeurIPS25_latent-mm-cot, Wang_NeurIPS25_unified-mm-cot, Cheng_NeurIPS25_visual-thoughts-mm-cot}.
Despite improving final prediction accuracy, CoT methods often suffer from the ignorance of the intermediate rationales during next-token prediction.
Consequently, they lack a strict causal dependence on the rationale content, resulting in limited interpretability~\cite{Jie_NAACL24_interpretability_cot, Barez_2025_cot_is_not_explainability, Yamaguchi_2025_red-mm-cot}.
A recent study~\cite{Yamaguchi_2025_red-mm-cot} attempted to increase reliance on rationale by ensembling image-conditional and rationale-conditional distributions at decoding time.
However, since the final outputs are still conditioned on the image input, LVLMs still lack a structural guarantee of causal dependency between the rationales and final predictions.
In this context, our RBMs introduce an inherently interpretable framework for LVLMs, in which the decoding method enforces a strict causal dependency between the rationales and final predictions by constructing next-token distributions from ensembles of rationale-conditional distributions, without images.

\section{Preliminaries}\label{sec:preliminaries}
This section introduces the formalization of next-token prediction in LVLMs and multi-modal CoT prompting principles as the basis for our proposed method.

\subsection{Next Token Prediction by LVLMs}
Consider an autoregressive LVLM, parameterized by $\theta$, which is trained on large-scale image-text datasets to process images as input for its backbone LLM.
Given an input image $x$ and query $q$, an LVLM generates an output token sequence $\bm{y}=(y_1,\dots,y_L)\in\mathcal{V}^L$ following:
\begin{equation}
     p(\bm{y}|x,q) = \prod^{L}_{i=1}p_\theta(y_i|\bm{y}_{<i},x,q),\label{eq:lvlm_cond_dist}
\end{equation}
where $L$ is the token length, $\mathcal{V}$ is the token vocabulary, and $\bm{y}_{<i}$ denotes the preceding output tokens.
As in autoregressive LLMs, $p_\theta(y_i|\bm{y}_{<i},x,q)$ over $\mathcal{V}$ is obtained by the softmax operation to the model's output $\operatorname{logits}_\theta(y_i|\bm{y}_{<i},x,q)$:
\begin{gather}
     p_\theta(y_i|\bm{y}_{<i},x,q) = \operatorname{softmax}(\operatorname{logits}_\theta(y_i|\bm{y}_{<i},x,q)) \\
     = \frac{\exp(\operatorname{logits}_\theta(y_i|\bm{y}_{<i},x,q))}{\sum_{{w}\in\mathcal{V}}\exp(\operatorname{logits}_\theta(w|\bm{y}_{<i},x,q))}.\label{eq:lvlm_softmax}
\end{gather}
Each token $y_i$ is generated from $p_\theta(y_i|\bm{y}_{<i},x,q)$ via a decoding strategy such as greedy decoding, i.e., $y_i = \argmax_{w\in\mathcal{V}}p_\theta(w|\bm{y}_{<i},x,q)$.
We define $\operatorname{generate(\cdot)}$ as a utility function for an arbitrary decoding strategy (e.g., greedy decoding):
\begin{equation}
     \bm{y} = \operatorname{generate}_\theta(x,q).\label{eq:lvlm_generate}
\end{equation}

\subsection{Multi-modal Chain-of-Thought}\label{sec:mm-cot}
Similar to text-modal CoT reasoning~\cite{Wei_NeurIPS22_CoT, Kojima_NeurIPS22_lets_think_step_by_step}, multi-modal CoT aims to enhance LVLM reasoning by incorporating intermediate rationale generation during decoding~\cite{Zhang_TMLR_mm-cot, Mondal_AAAI24_kam-cot, Mitra_CVPR24_CCoT, Tang_CVPR25_reason-before-retrieve-mm-cot}.
Multi-modal CoT typically consists of two reasoning steps: (i) rationale generation and (ii) output generation.
First, LVLMs generate a rationale $r$ from an input image $x$, query prompt $q$, and instruction $I_r$, e.g., ``\texttt{\small Given the image, generate the reasoning as a rationale for answering the question}'', as follows:
\begin{equation}
     r = \operatorname{generate}_\theta(x,I_r,q).\label{eq:lvlm_rationale_generation}
\end{equation}
Previous work has proposed several forms of $r$, such as free-form reasoning texts~\cite{Zhang_TMLR_mm-cot} and schema-less scene graphs in JSON format~\cite{Mitra_CVPR24_CCoT}.
Then, LVLMs output the final answer by next-token prediction:
\begin{align}
     \bm{y} = \operatorname{generate}_\theta(x,r,q).\label{eq:lvlm_cot_generate}
\end{align}
While this bi-level prediction improves the final output accuracy, it does not guarantee that the output is grounded solely in the generated rationale $r$ because the input context in Eq.~\eqref{eq:lvlm_cot_generate} still includes the image $x$.
In this regard, standard CoT methodologies cannot provide rigorous interpretability based on causal dependencies between rationales and outputs~\cite{Barez_2025_cot_is_not_explainability}.
In fact, Yamaguchi~et.al.~\cite{Yamaguchi_2025_red-mm-cot} empirically showed that the next token predictions in LVLMs often ignore the rationale and instead focus on visual tokens of input images within the attention modules.
A naive alternative would be to condition the generation solely on the rationale, i.e., $\operatorname{generate}_\theta(r,q)$. 
However, this still lacks structured interpretability of each rationale component and its respective contribution score, unlike CBMs defined by Eq.~\eqref{eq:cbm_obj}.
In this work, we aim to transform LVLMs' multi-modal CoT to be inherently interpretable without compromising their zero-shot task performance.
We achieve this by introducing structured rationales that explicitly represent semantic relationships between visual concepts with a decoding strategy that enforces the causal dependency between rationales and outputs in a plug-and-play manner.

\begin{figure*}[tbp]
    \centering
    \includegraphics[width=0.9\linewidth]{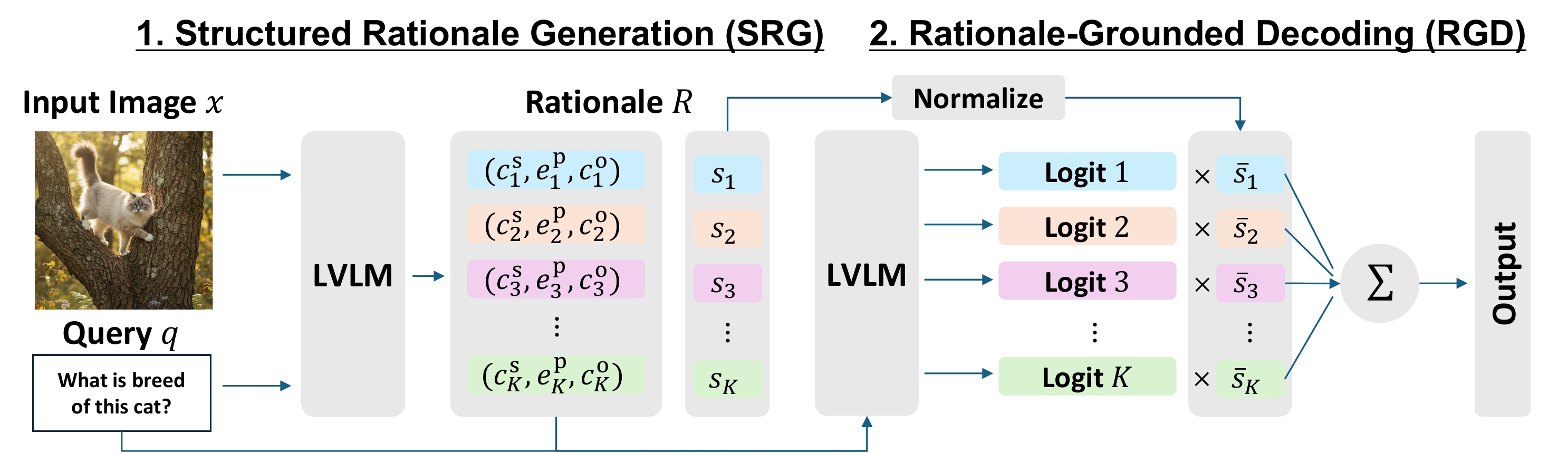}
    \caption{
    \textbf{Proposed Method.} Relational Bottleneck Models (RBMs) consisting of (i) Structured Rationale Generation (SRG) and (ii) Rationale-Grounded Decoding (RGD).
    (i) SRG produces rationales for the final output by a set of semantic relation triplets of visual concepts in an input image.
    Each rationale is defined by a semantic relational triplet (subject, predicate, object) and an importance score.
    This structured rationale representation enables rich interpretability of high-level visual semantic relationships between concepts beyond the independent visual concept rationales often used in CBMs.
    (ii) Given rationales from an input image and a query, RGD constructs a composite next-token distribution by ensembling conditional distributions for each semantic relation triplet via importance-weighted product-of-experts.
    By decoding output tokens with this composite distribution, RBMs ensure the interpretability of the final outputs.
    }
    \label{fig:method}
    \vspace{-3mm}
\end{figure*}

\section{Method}\label{sec:method}
We propose Relational Bottleneck Models (RBMs), a family of inherently interpretable models built on top of LVLMs in a plug-and-play manner.
As illustrated in Fig.~\ref{fig:method}, RBMs consist of two core components: (i) Structured Rationale Generation (SRG) and (ii) Rationale-Grounded Decoding (RGD) for the bi-level generations of rationales and final outputs, yielding a causal dependency between them.
The rest of this section describes the details of SRG, RGD, and the inference algorithm of RBMs shown in Algorithm~\ref{alg:inference}.

\begin{figure}[t]
\vspace{-3mm}
\begin{algorithm}[H]
    \caption{Inference of Relational Bottleneck Models}\label{alg:inference}
    \begin{algorithmic}[1]
        \Require{Input image $x$, query $q$, LVLM parameterized by $\theta$, hyper-parameter $\alpha$}
        \Ensure{Decoded output sequence $\bm{y}$}
        \State $R \gets \text{generate}_\theta(x, I_R, q)$ \Comment{{\footnotesize \color{gray}Structured Rationale Generation}}
        \State $\bm{y} \gets \emptyset$        
        \While{$|\bm{y}| < L$} \Comment{{\footnotesize\color{gray} Loop for Rationale-Grounded Decoding}}
            \State $\hat{\operatorname{logits}}_\theta(y_i|\bm{y}_{<i},R,q) \gets$ Eq.~\eqref{eq:grd_logits}
            \State $y \gets \argmax_{w\in\mathcal{V}}~\text{softmax}(\hat{\operatorname{logits}}_\theta(w|\bm{y}_{<i},R,q))$
            \State $\bm{y}.\text{append}(y)$ \Comment{{\footnotesize\color{gray}Add last token to output sequence}}
        \EndWhile
        
        \State \Return $\bm{y}$
    \end{algorithmic}
\end{algorithm}
\vspace{-5mm}
\end{figure}

\subsection{Structured Rationale Generation (SRG)}
In the first step, an RBM generates a set of $K$ rationales $R$ for an input image $x$.
Each rationale is defined by a tuple of a visual semantic relation triplet $r_k$ and an importance score $s_k$, which quantifies its contribution to the final output:
\begin{align}
    R := \{(r_k, s_k) | r_k = (c^\mathrm{s}_k, e^{\mathrm{p}}_k, c^\mathrm{o}_k), s_k\in[0,1]\}^K_{k=1},
\end{align}
where $c^\mathrm{s}_k$ is the subject concept, $e^\mathrm{p}_k$ is the predicate, and $c^\mathrm{o}_k$ is the object concept.
The triplet $r_k$ explicitly models a visual semantic relationship between two concepts in the image, e.g., (``A cat'', ``stands on'', ``moss-covered branch'').
By connecting $c^\mathrm{s}_k$ and $c^\mathrm{o}_k$ via $e^\mathrm{p}_k$, SRG provides an explicit rationale for the final prediction, whereas existing CBMs typically isolate individual concepts like ``A Cat'' and ``moss-covered branch.''

Meanwhile, $s_k$ represents the relative contribution of the $k$-th rationale to the final prediction.
Prior work on CBMs often computes concept contribution scores based on prediction probabilities from trained concept predictors~\cite{Koh_ICML20_concept_bottleneck, Ramaswamy_CVPR23_overlooked_factors_CBMs}.
In contrast, we cannot directly compute these scores since RBMs operate without any additional training or specialized evaluation modules for rationales, an intentional trade-off to preserve the zero-shot capability and plug-and-play nature of RBMs.
For the same reason, we avoid using attention scores~\cite{Kobayashi_EMNLP20_attention_contributions, Basu_NeurIPS24_understanding_lvlms_attention, Kang_ICLR25_visual_attention_sink}, which necessitate architecture-dependent configurations.
Instead, we propose leveraging the LVLM's self-verbalized rationale importance scores to represent $s_k$.
While such self-verbalized scores are often miscalibrated~\cite{Groot_2024_overconfidence_calibration-lvlms, Xuan_EMNLP25_seeing_is_believing_calibration-lvlms}, our objective is merely to obtain a relative proxy for importance to ground the final predictions, rather than perfectly calibrated probabilities.
Thus, we treat these verbalized values as the model's intrinsic importance scores and faithfully reflect them during decoding.
Given the self-verbalized importance scores $\{s_k\}^K_{k=1}$, we compute the normalized scores via a softmax function:
\begin{align}
    \bar{s}_k = \frac{\exp(s_k)}{\sum^{K}_{j=1}\exp(s_j)}.
\end{align}

\subsection{Rationale-Grounded Decoding (RGD)}
To ensure a strict causal dependency between intermediate rationales and final predictions, RBMs compute the next-token distributions for final decoding by aggregating multiple rationale-conditional distributions $p_\theta(y_i|\bm{y}_{<i},r_k, q)$, while strictly discarding the input image from the context.
To formalize this idea, we propose RGD, which constructs the next-token distribution by a weighted product-of-experts (PoE,~\cite{Hinton2002_poe}) ensemble of the rationale-conditional distributions.
Given a generated rationale $R$, the next-token distribution in RGD is defined as follows:
\begin{gather}
     \hat{p}_\theta(y_i|R,q) \propto \prod^K_{k=1} p_\theta(y_i | \bm{y}_{<i}, r_k, q)^{\bar{s}_k}.\label{eq:rgd}
\end{gather}
Since the PoE distribution in Eq.~\eqref{eq:rgd} acts as an AND operation across the rationale-conditional distributions, it effectively generates tokens that are consistently supported by SRG.
Furthermore, raising the rationale-conditional probabilities to the power of $\bar{s}_k$ ensures that the predictions rely more heavily on rationales with higher self-reported importance, improving the interpretability and faithfulness.

\begin{figure}[tbp]
\centering
\begin{lstlisting}[language=json, basicstyle=\scriptsize\ttfamily]
{
  "type": "object",
  "properties": {
    "relations": {
      "type": "array",
      "items": {
        "type": "object",
        "properties": {
          "subject": { "type": "string" },
          "predicate": { "type": "string" },
          "object": { "type": "string" },
          "importance_score": {
            "type": "number",
            "minimum": 0,
            "maximum": 1
          }
        },
        "required": ["subject", "predicate", "object", "importance_score"]
      }
    }
  }
}
\end{lstlisting}
\caption{JSON Schema of Semantic Relation Triplets in Structured Rationale Generation}
\label{fig:srg_json_schema}
\vspace{-5mm}
\end{figure}

\subsection{Algorithm and Implementation}
Here, we describe the details of the RBM inference algorithm and its implementations.
The overall algorithm is defined in Algorithm~\ref{alg:inference}.

\noindent\textbf{Rationale Generation.}
As defined in Algorithm~\ref{alg:inference}, the first step generates up to $K$ structured rationales from the input image $x$:
\begin{equation}
     R = \operatorname{generate}_\theta(x,I_R,q),\label{eq:srg_generation}
\end{equation}
where $I_R$ is the instruction for SRG, which specifies the output format in JSON:
\begin{promptbox}[title={Instruction for Structured Rationale Generation $I_R$}]
\# Instruction

Analyze the image and identify key visual relations with a relative importance score to answer the question.
Infer the relative importance scores for each rationale based on the contribution to constructing the final answer.
Provide your response in the following JSON Schema:

... (See Fig.~\ref{fig:srg_json_schema}) ...

\# Response

[Identify up to $K$ most relevant relations here.]
\end{promptbox}
\noindent We used JSON Schema~\cite{Pezoa_2016_foundations_json-schema} to rigorously define the output specifications as described in Fig.~\ref{fig:srg_json_schema}.

\noindent\textbf{Decoding for Final Output.}
From lines 3 to 7 in Algorithm~\ref{alg:inference}, we decode tokens via RGD.\@
To mathematically satisfy the PoE formulation, we compute the next token distribution with an LVLM parameterized by $\theta$ similar to Eq.~\eqref{eq:lvlm_softmax}:
\begin{gather}
     \hat{p}_\theta(y_i|\bm{y}_{<i},R,q) = \operatorname{softmax}(\hat{\operatorname{logits}}_\theta(y_i|\bm{y}_{<i},R,q)), \label{eq:red_prob}\\
     \hat{\operatorname{logits}}_\theta(y_i|\bm{y}_{<i},R,q) = \sum^K_{k=1} \bar{s}_k \operatorname{logits}_\theta(y_i|\bm{y}_{<i},r_k,q),\label{eq:grd_logits}
\end{gather}
We compute these aggregated logits in Eq.~\eqref{eq:grd_logits} using batched parallelization across the $K$ rationales on a GPU.

\section{Experiments}\label{sec:experiments}

\begin{table}[t]
\centering
\caption{
\textbf{Rationale Accuracy}.
Quantitative evaluation of absolute rationale quality based on Faithfulness, Query-Relevance, and Coverage on a 1-5 scale. RBMs significantly outperform LVLM-CBMs, particularly in Query-Relevance, verifying that they provide precise and contextually appropriate evidence directly linked to the query.
}
\label{tb:rationale_accuracy}
\resizebox{\linewidth}{!}{
\begin{tabular}{@{}lcccc|cccc}
\toprule
 & \multicolumn{4}{c}{\textbf{GQA}} & \multicolumn{4}{c}{\textbf{MMMU-Pro}} \\
\cmidrule(r){2-5}\cmidrule(r){6-9} 
 & Faith. & Query-Rel. & Coverage & Avg. & Faith. & Query-Rel. & Coverage & Avg. \\
\midrule
\multicolumn{9}{@{}l}{\cellcolor{LightRed}\textbf{Qwen3-VL-8B}} \\
LVLM-CBM & 2.42 & 1.08 & 1.44 & 1.65 & 1.79 & 1.26 & 1.31 & 1.45 \\
RBM & \textbf{3.52} & \textbf{3.52} & \textbf{3.48} & \textbf{3.51} & \textbf{3.10} & \textbf{3.62} & \textbf{3.10} & \textbf{3.27} \\
\midrule
\multicolumn{9}{@{}l}{\cellcolor{LightBlue}\textbf{Gemma3-12B}} \\
LVLM-CBM & 1.96 & 1.03 & 1.04 & 1.34 & 1.99 & 1.63 & 1.43 & 1.68  \\
RBM  & \textbf{3.61} & \textbf{3.38} & \textbf{3.35} & \textbf{3.45} & \textbf{2.67} & \textbf{3.35} & \textbf{2.87} & \textbf{2.96} \\
\midrule
\multicolumn{9}{@{}l}{\cellcolor{LightYellow}\textbf{InternVL3.5-8B}} \\
LVLM-CBM & 2.33 & 1.14 & 1.39 & 1.62 & 1.69 & 1.31 & 1.16 & 1.39 \\
RBM  & \textbf{3.22} & \textbf{3.13} & \textbf{3.09} & \textbf{3.15} & \textbf{2.49} & \textbf{2.91} & \textbf{2.32} & \textbf{2.57} \\
\bottomrule
\end{tabular}
}
\vspace{-5mm}
\end{table}

\subsection{Settings}
\noindent\textbf{Models.}
We used public LVLMs on HuggingFace: Qwen3-VL~\cite{Bai_2025_qwen3-vl}, Gemma3~\cite{Google_2025_gemma3}, and InternVL3.5~\cite{Wang_2025_internvl3.5}.

\noindent\textbf{Baselines.}
Since there is no existing zero-shot inherent interpretable LVLM, we constructed \textit{LVLM-CBM}, which performs bi-level prediction (image-to-concept, then concept-to-output) strictly excluding the input image from the final decoding context.
Specifically, we generate rationale concepts $R^\mathrm{c}=\{c_k\}^K_{k=1}$ and then predict final outputs by $p_\theta(y_i|\bm{y}_{<i},R^\mathrm{c},q)$.
As the non-interpretable baseline, we used multi-modal \textit{CoT} with text reasoning rationales~\cite{Zhang_TMLR_mm-cot} that directly decode outputs conditioned on both the rationale and the input image by $p_\theta(y_i|\bm{y}_{<i},x,r,q)$.

\noindent\textbf{Benchmarks.}
We used vision-language benchmarks in various tasks and domains: \textit{GQA}~\cite{Hudson_CVPR19_gqa}, \textit{TextVQA}~\cite{Singh_CVPR19_textvqa}, \textit{MME}~\cite{Fu_2023_mme}, \textit{COCO Caption}~\cite{Lin_2014_COCO}, \textit{MMMU}~\cite{Yue_CVPR24_mmmu}, and \textit{MMMU-Pro}~\cite{Yue_ACL25_mmmu-pro}; see the Appendix for details.

\noindent\textbf{Evaluation Metrics.}
For \textit{Task Performance}, we report exact matching accuracy with the ground-truth labels for GQA, TextVQA, and MMMU/MMMU-Pro, and the CIDEr score~\cite{Vedantam_CVPR15_cider} for COCO Caption.
For interpretability, we used relative and absolute metrics based on LLM-as-a-Judge~\cite{Gu_2024_llm-as-a-judge-survey, Chen_ICML24_mllm-as-a-judge} using GPT-5-mini~\cite{Singh_2025_openai-gpt-5}.
For the absolute metrics, we report three five-point scale scores (1-5) regarding rationale accuracy: 
\textit{Faithfulness}, whether the rationale accurately describes visual features causally depending on the answer or hallucinates them,
\textit{Query-Relevance}, whether the rationale directly enables deriving the answer, or it is unrelated/insufficient to the query,
\textit{Coverage}, whether the rationale comprehensively identifies the key evidence needed to answer the question, or lacks important evidence.
For the relative metric, \textit{Rationale Preference} is a score equal to the win rate of target models when the LLM evaluator compares the rationales of target and competitor models in terms of interpretability.
We provide more details in the Appendix.
To assess the causal dependency between rationales and final outputs, we perform \textit{Intervention} analysis by replacing generated rationales with high-quality ones from Gemini3-Flash~\cite{google_2025_gemini3} or randomly sampled ones from the high-quality rationales.\looseness-1

\noindent\textbf{Inference Protocol.}
We used greedy decoding to determine the next token for each iteration, with a maximum new token length of 1024, ensuring that all outputs are produced deterministically.
Unless specified, we set $K$ to 5 for RBMs and 10 for LVLM-CBMs.
We used an NVIDIA H100 with 80GB of memory.\looseness-1

\begin{figure}[t]
    \centering
    \begin{minipage}[t]{0.32\linewidth}
      \includegraphics[width=\columnwidth]{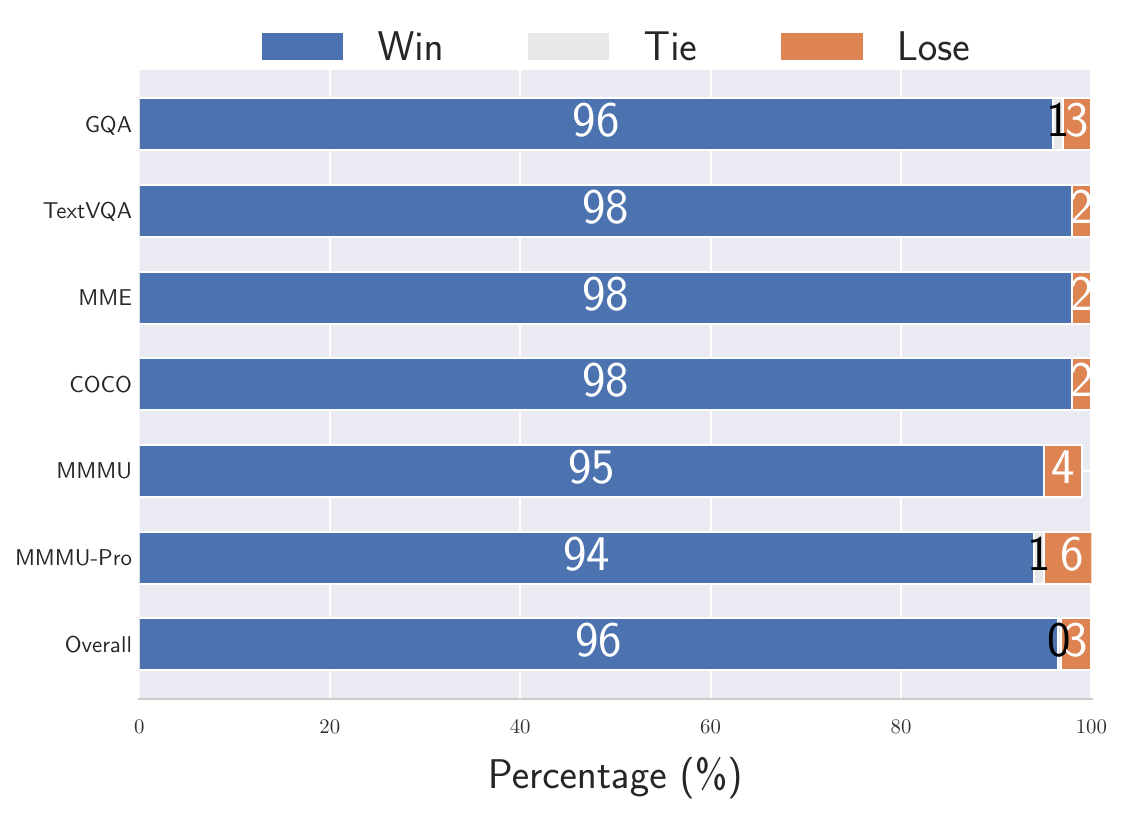}
      \subcaption{Qwen3-VL-8B}
    \end{minipage}
    \begin{minipage}[t]{0.32\linewidth}
      \includegraphics[width=\columnwidth]{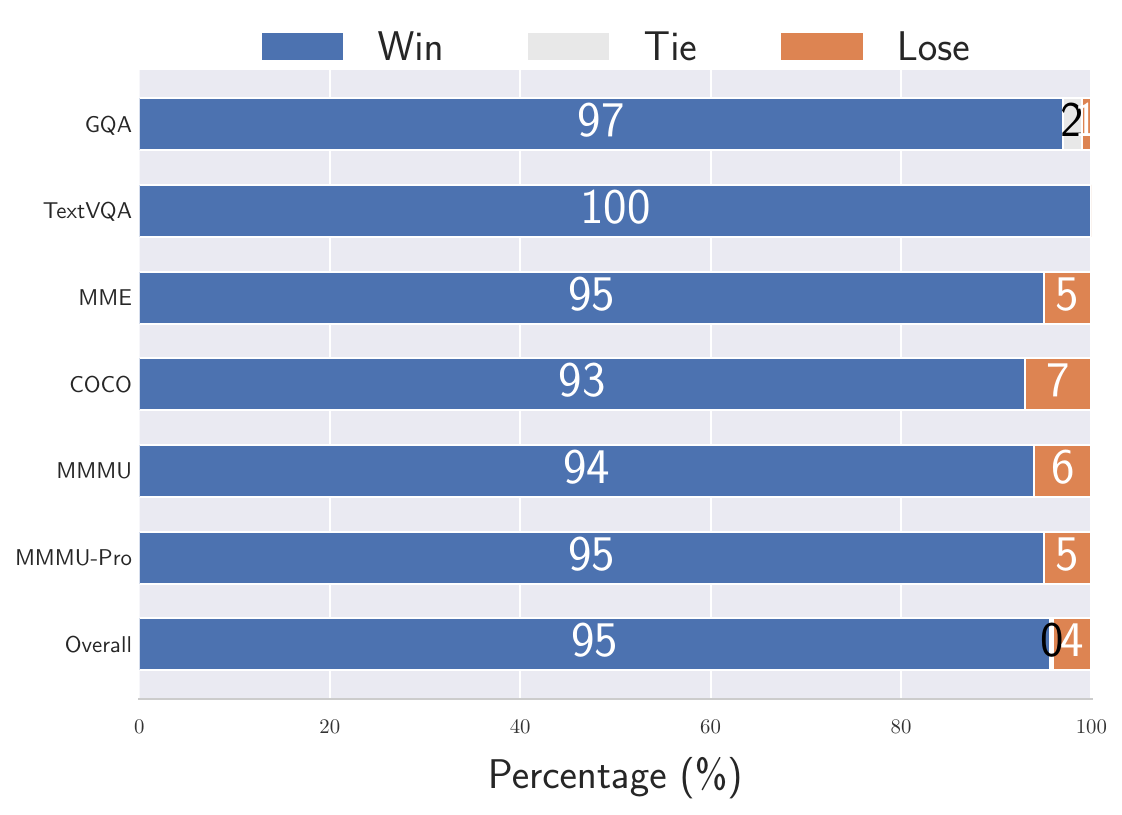}
      \subcaption{Gemma3-12B}
    \end{minipage}
    \begin{minipage}[t]{0.32\linewidth}
      \includegraphics[width=\columnwidth]{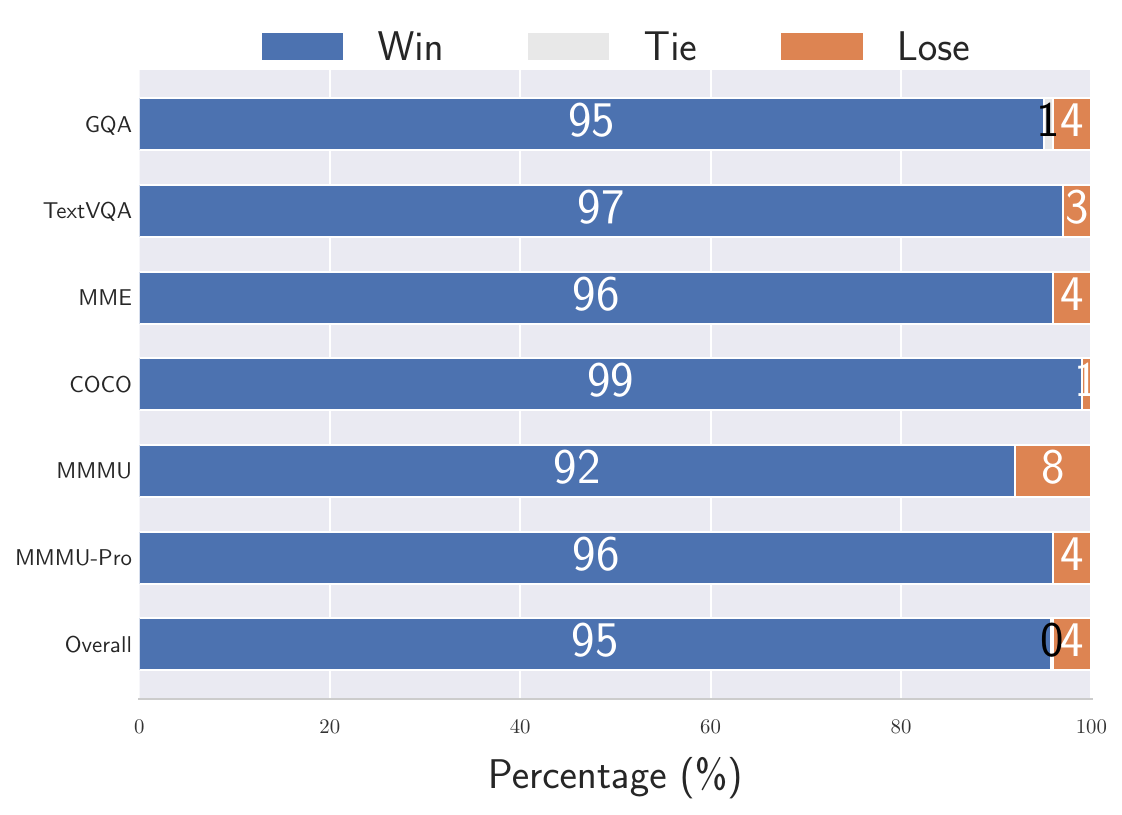}
      \subcaption{InternVL3.5-8B}
    \end{minipage}    
    \vspace{-3mm}
    \caption{
        \textbf{
        Rationale Preference Evaluation (RBM vs. LVLM-CBM)}. 
        A relative comparison of rationale quality generated by RBMs and LVLM-CBMs using GPT-5-mini as a judge. RBMs achieve a win rate exceeding 90\% across all cases, demonstrating their superior utility in generating highly interpretable reasoning.
    }
    \label{fig:rationale_preference}
    \vspace{-5mm}
\end{figure}

\subsection{Evaluations on Interpretability}
\noindent\textbf{Rationale Accuracy.}
Table~\ref{tb:rationale_accuracy} shows the rationale accuracy of LVLM-CBM and RBM.\@
RBMs outperformed LVLM-CBMs in all cases.
For Faithfulness, the gain indicates that RBM's RGD can improve the causal dependency between rationales and final outputs as expected.
RBMs provided large improvements in Query-Relevance, indicating how closely the generated rationales align with the query.
For Coverage, although LVLM-CBMs ($K=10$) can generate twice the maximum number of rationales as RBMs ($K=5$), RBMs achieved much higher scores, indicating that RBMs offer better interpretability by explicitly representing visual semantic relations, not just increasing concepts.
These results demonstrate that RBMs' structured rationales can more accurately and directly express information relevant to the query and answer by providing semantic relationships of visual concepts, beyond providing isolated concepts in CBMs.

\noindent\textbf{Rationale Preference.}
Fig.~\ref{fig:rationale_preference} compares the rationales of RBMs and LVLM-CBMs in the format of win (RBM rationale is superior), tie (neither rationale is superior), and lose (LVLM-CBM rationale is superior) on 100 randomly sampled instances from each benchmark.
RBMs achieved a win rate exceeding 90\% across all cases, demonstrating their superior utility for interpreting outputs.

\begin{table}[t]
\centering
\caption{\textbf{Intervention Analysis with High-Quality (Gemini3) and Irrelevant (Random) Rationales}.
The significant shifts in accuracy confirm that RBMs are strictly grounded in their rationales.
}
\label{tb:intervention}
\vspace{-2mm}
\begin{tabular*}{\columnwidth}{@{\extracolsep{\fill}}lccc@{}}
\toprule    
 & \multicolumn{3}{c}{\textbf{GQA}} \\
 & Self ($\uparrow$) & Gemini3 ($\uparrow$) & Random ($\downarrow$) \\
\midrule
\multicolumn{4}{@{}l}{\cellcolor{LightRed}\textbf{Qwen3-VL-8B}} \\
CoT & 51.96 & 55.73 & 26.10 \\\midrule
LVLM-CBM & 46.73 & 48.16 & 24.06 \\
RBM & \textbf{54.98} & \textbf{59.34} & \textbf{12.39} \\
\midrule
\multicolumn{4}{@{}l}{\cellcolor{LightBlue}\textbf{Gemma3-12B}} \\
CoT & 24.64 & 38.78 & 4.50  \\\midrule
LVLM-CBM & 12.61 & 20.51 & 0.80 \\
RBM & \textbf{17.42} & \textbf{21.29} & \textbf{0.58} \\
\midrule
\multicolumn{4}{@{}l}{\cellcolor{LightYellow}\textbf{InternVL3.5-8B}} \\
CoT & 54.56 & 56.81 & 32.64 \\\midrule
LVLM-CBM & 45.76 & 46.95 & 21.55 \\
RBM & \textbf{48.03} & \textbf{48.82} & \textbf{11.80} \\
\bottomrule
\end{tabular*}
\vspace{-5mm}
\end{table}
\begin{figure}[t]
\centering
\includegraphics[width=\columnwidth]{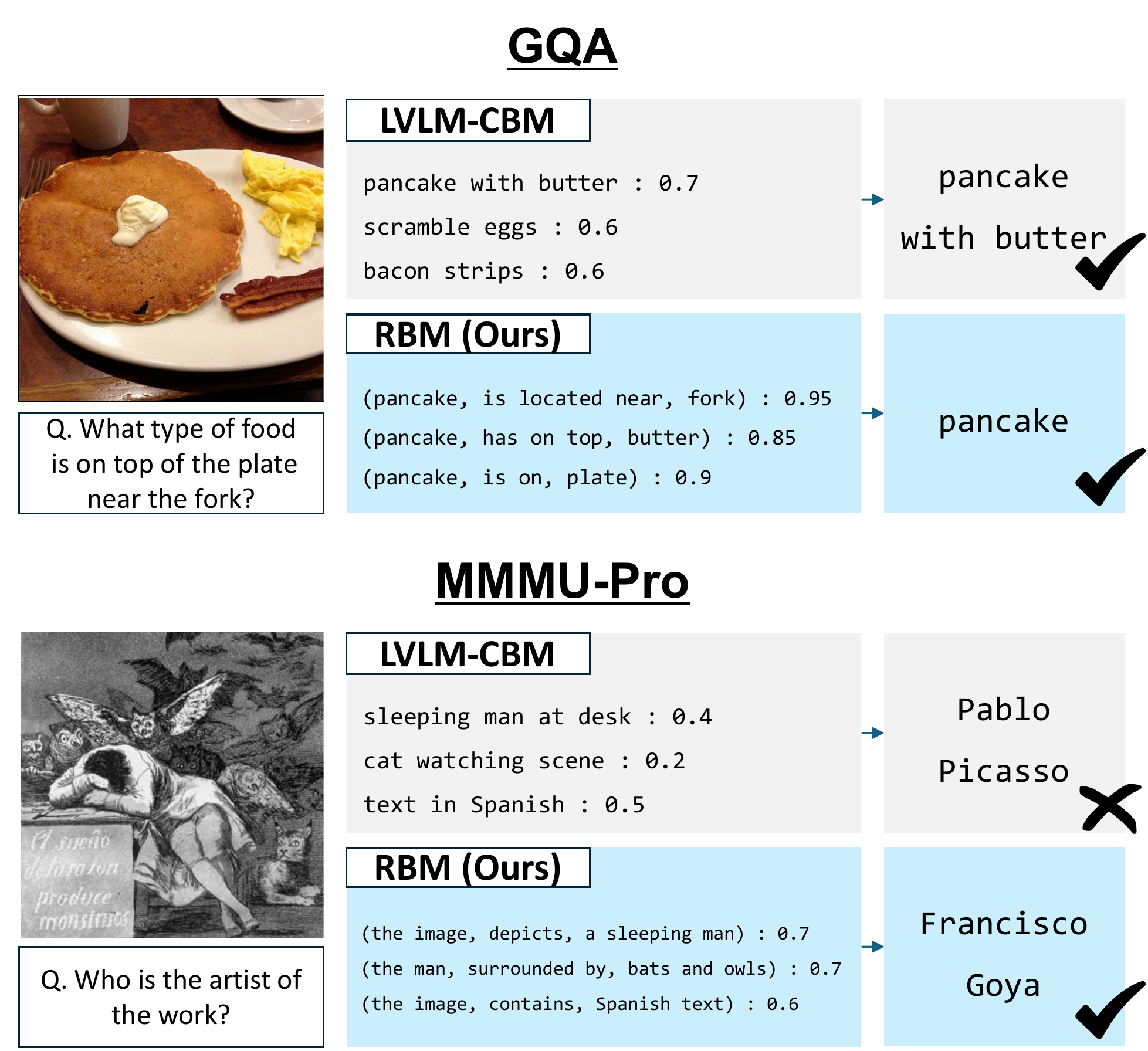}
\captionof{figure}{
\textbf{Qualitative Evaluation}.
Unlike CBMs that output isolated concepts, RBMs successfully extract structured semantic relations, yielding clearer and more logical reasoning paths.
}
\label{fig:qualitative}
\vspace{-3mm}
\end{figure}

\begin{figure}[t]
    \centering
    \includegraphics[width=\linewidth]{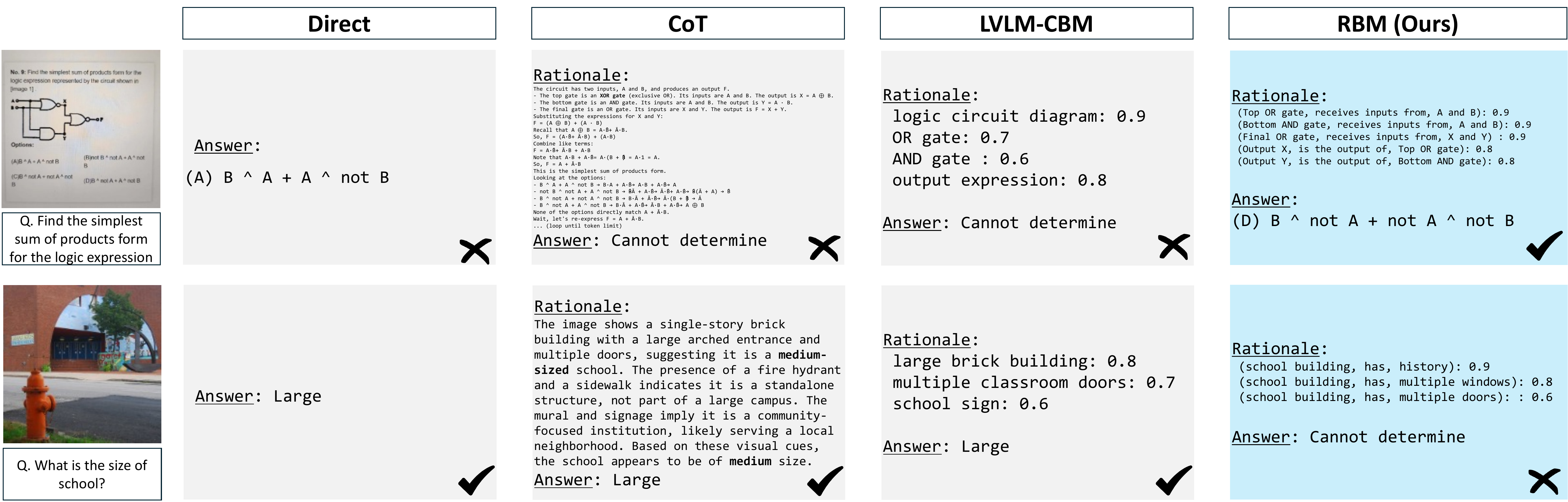}
    \vspace{-4mm}
    \caption{
    \textbf{Detailed Visualization}.
    RBMs can strongly perform on reasoning tasks that require extracting and integrating semantic relationships from the overall image (top), while they may fail on perceptual queries that can be answered without logical reasoning.
    }
    \label{fig:viz2}
    \vspace{-5mm}
\end{figure}

\noindent\textbf{Intervention Analysis.}
To validate the rationale-output causal dependency, we swap the generated rationales with positive (Gemini 3) and negative (random) interventions as shown in Table~\ref{tb:intervention}. 
While CoT shows low sensitivity to negative interventions (indicating weak rationale reliance), RBMs exhibit dramatic output shifts for both interventions. 
This strong causal grounding confirms that RBMs strictly base their final predictions on the provided rationales, surpassing both CoT and LVLM-CBMs.

\noindent\textbf{Qualitative Analysis.}
Fig.~\ref{fig:qualitative} visualizes the output examples of LVLM-CBMs and RBMs; we used Qwen3-VL-8B and randomly selected examples in this experiment.
Unlike LVLM-CBMs, which list isolated visual concepts, RBMs generate more detailed semantic relationships that are closely tied to the input images and queries (e.g., ``fork'' and ``plate'' in GQA).
In the MMMU-Pro example, CBM attempts to answer by listing a single concept, whereas RBM gathers the information necessary to identify the artist of the painting through clear and logical relationships, offering users more interpretable rationales for final predictions.
We also conducted a detailed qualitative analysis as shown in Fig.~\ref{fig:viz2}.
While all baselines failed to yield the correct answer, RBMs correctly responded by extracting and integrating visual evidence across the entire image (Fig.~\ref{fig:viz2}, top).
As a failure case, the bottom part shows that RBMs may fail on perceptual queries that can be answered directly without reasoning, whereas simpler rationales from CBMs succeed, and CoT generates a correct answer by ignoring the rationale.
These results highlight that RBMs have an advantage in more complex visual reasoning tasks where humans require interpretable explanations for their answers.\looseness-1

\begin{table}[t]
\centering
\caption{
\textbf{Zero-shot Performance on Multiple Multi-modal Benchmarks}.
RBMs consistently surpass the inherently interpretable LVLM-CBMs while effectively narrowing the performance gap with non-interpretable methods (Direct decoding and CoT).
}
\label{tb:zero-shot_perf}
\resizebox{\linewidth}{!}{
\begin{tabular}{@{}lcccccc}
\toprule
 & \textbf{GQA} & \textbf{TextVQA} & \textbf{MME} & \textbf{COCO} & \textbf{MMMU} & \textbf{MMMU-Pro} \\
\midrule
\multicolumn{7}{@{}l}{\cellcolor{LightRed}\textbf{Qwen3-VL-8B}} \\
Direct & 60.73 & 78.34 & 89.30 & 21.02 & 42.00 & 23.29 \\
CoT & 51.96 & 60.80 & 67.02 & 3.08 & 33.11 & 15.66  \\\midrule
LVLM-CBM & 46.73 &	63.28 &	\textbf{69.47} & 14.43 & 35.33 &	14.28 \\
RBM  & \textbf{54.98} &	\textbf{68.76} & 69.00 & \textbf{20.83} & \textbf{37.78} & \textbf{15.14}\\
\midrule
\multicolumn{7}{@{}l}{\cellcolor{LightBlue}\textbf{Gemma3-12B}} \\
Direct & 44.17 & 76.08 & 84.16 & 33.14 & 38.33 & 17.86\\
CoT & 24.64 & 71.82 & 85.80 & 19.49 & 45.11 & 20.58\\\midrule
LVLM-CBM & 12.61 &	56.70 &	62.64 &	9.10 &	36.56 &	18.55\\
RBM  & \textbf{17.42} & \textbf{60.40} & \textbf{64.03} & \textbf{18.94} & \textbf{44.56} & \textbf{20.35}\\
\midrule
\multicolumn{7}{@{}l}{\cellcolor{LightYellow}\textbf{InternVL3.5-8B}} \\
Direct & 61.41 & 78.34 & 87.19 & 52.93 & 47.56 & 22.83\\
CoT & 54.56 & 74.14 & 92.92 & 51.70 & 34.11 & 14.68 \\\midrule
LVLM-CBM & 45.76 &	54.04 &	65.63 &	10.32 &	\textbf{39.78} & \textbf{16.88}\\
RBM  & \textbf{48.03} & \textbf{69.44} & \textbf{71.32} & \textbf{18.96} & 38.56 &	15.90\\
\bottomrule
\end{tabular}
}
\vspace{-3mm}
\end{table}

\subsection{Evaluations on Zero-shot Performance}
Table~\ref{tb:zero-shot_perf} shows the zero-shot performance of RBMs compared with direct decoding, CoT, and LVLM-CBMs across diverse multi-modal benchmarks.
Overall, RBMs consistently achieve higher average accuracy than LVLM-CBMs, the inherently interpretable baseline, across all three LVLM backbones.
This suggests that encoding rationales as SRG captures more actionable contextual information than isolated concepts.

However, we observed an expected performance drop compared to non-interpretable methods (Direct and CoT) on tasks requiring fine-grained visual details, such as COCO Caption.
This highlights the inherent trade-off of strict rationale-bottleneck architectures: filtering out low-level visual information guarantees causal faithfulness but sacrifices descriptive capacity.
Nevertheless, RBMs successfully mitigate this gap better than existing CBM approaches, maintaining competitive zero-shot performance.

\begin{table}[t]
\centering
\caption{
\textbf{Ablation Study}.
Removing SRG, RGD, or self-reported importance scores ($\bar{s}_k$) degrades task accuracy and/or interpretability, confirming the design validity. 
}
\label{tb:ablation}
\resizebox{0.95\columnwidth}{!}{
\begin{tabular}{@{}lccccc@{}}
\toprule    
 & \multicolumn{5}{c}{\textbf{GQA-Val}} \\
 & Task Acc{.} & Faith{.} & Query-Rel{.} & Coverage & Avg{.}~Rationale Acc{.} \\
\midrule
\multicolumn{6}{@{}l}{\cellcolor{LightRed}\textbf{Qwen3-VL-8B}} \\
RBM & 53.0 & 3.52 & 3.52 & 3.48 & 3.51 \\
~~w/o SRG w/ Concept Rationale & 49.0 & 2.72 & 2.23 & 2.06 & 2.34 \\
~~w/o RGD & 59.0 & 3.05 & 3.52 & 3.48 & 3.35 \\
~~w/ Uniform Weights ($\bar{s}_k=1$) & 38.0 & 3.39 & 3.52 & 3.48 & 3.46 \\
~~w/ Permuted Weights ($\bar{s}_k=\bar{s}_{j\neq k}$) & 46.0 & 3.31 & 3.52 & 3.48 & 3.44 \\
\bottomrule
\end{tabular}
}
\vspace{-5mm}
\end{table}


\subsection{Ablation Study}
In this section, we provide a detailed analysis of varying the component selections of RBMs.
We used the GQA-Val dataset, a subset of the GQA training set containing 100 randomly sampled instances.
Table~\ref{tb:ablation} shows the ablation results of RBMs.
Here, we replace 
(i) the rationale format in SRG with concepts in CBMs, i.e., $R=\{(c_k,s_k)\}^K_{k=1}$, 
(ii) the decoding strategy from RGD to naive conditional distribution, i.e., $p_\theta(y|R,q)$, 
(iii) the importance weights from self-reported weights to uniform weights, i.e., $\bar{s}_k=1$, 
(iv) the importance weights by a random permutation, i.e., $\bar{s}_k=\bar{s}_{j\neq k}$.
For (i), the results show that SRG improves both task accuracy and rationale interpretability in RBMs, indicating the necessity of SRG.
For (ii), while RBM w/o RGD can improve task performance, it degrades the faithfulness score and the entire interpretability.
This is because the naive conditional distribution $p_\theta(y|R,q)$ cannot guarantee the contributions of each rationale $r_k$ with the importance score, highlighting the better interpretability with explicit rationale-dependency by Eq.~\eqref{eq:rgd}.
For (iii) and (iv), we observed that the uniform and permuted weights largely degrade the task accuracy and the faithfulness, confirming that the self-reported importance is an essential factor in achieving both task performance and interpretability.
These ablations demonstrate the design validity of our method.

\subsection{Effects of $K$}\label{append:effects-of-K}
\begin{figure}[t]
\centering
\includegraphics[width=\columnwidth]{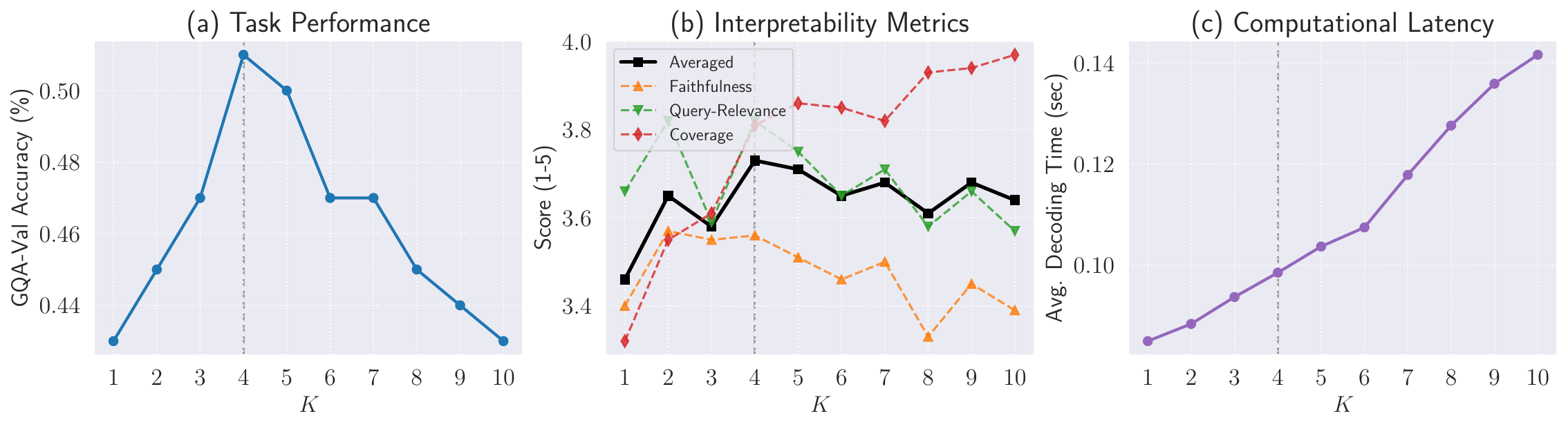}
\captionof{figure}{
\textbf{Effects of $K$ (Qwen3-VL-8B, GQA-Val)}.
A smaller $K$ (e.g., $K=4$) is optimal in practice, as a larger $K$ causes lower performance, less interpretability, and slower computations.
}
\label{fig:k_effects}
\vspace{-5mm}
\end{figure}

The maximum rationale number $K$ is a crucial factor determining the rationale contents and final outputs in RBMs.
We varied $K$ from 1 to 10 and measured task accuracy, interpretability, and latency as shown in Fig.~\ref{fig:k_effects}.
We used the GQA-Val dataset, a subset of the GQA training set containing 100 randomly sampled instances.
Notably, we observed that a larger $K$ does not yield higher task accuracy or interpretability.
This may be because a lot of rationales with a larger $K$ contain noisy rationales and thus confuse LVLMs in the decoding phase by PoE, which determines next-tokens by AND operations of each rationale-conditional distributions.
Besides, a larger $K$ incurs greater latency due to increased rationale generation latency.
These results indicate that selecting an appropriate $K$ is crucial for task performance, interpretability, and computational cost, and that in practice, a small $K$ (e.g., $K=4$ or $5$) is optimal.

\subsection{Model Scalability}\label{append:model-scalability}
\begin{figure}[tbp]
    \centering
    \includegraphics[width=\linewidth]{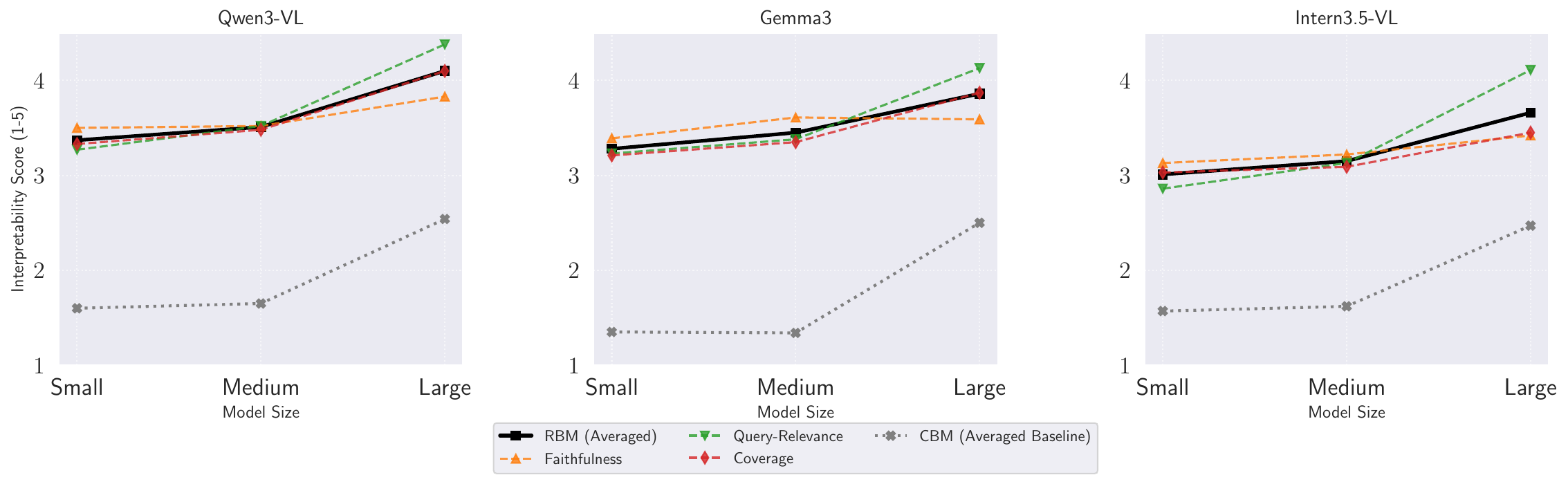}
    \caption{
    \textbf{Model Scalability in Interpretability}.
    }
    \label{fig:model_scale_interpretability}
    \vspace{-3mm}
\end{figure}

Here, we show the additional results when varying the model sizes of backbone LVLMs for RBMs.
We used the small, medium, and large models for each backbone, i.e., Qwen3-VL (4B, 8B, 32B), Gemma3 (4B, 12B, 27B), and InternVL3.5 (4B, 8B, 38B).
Table~\ref{tb:model_scale_task} and Fig.~\ref{fig:model_scale_interpretability} show the results of task performance and rationale accuracy (interpretability) on GQA.
We see that our RBMs achieved higher task accuracy and rationale accuracy than LVLM-CBMs for all models, indicating the scalability of RBMs for the model sizes.\looseness-1

\begin{table}[t]
\centering
\caption{
\textbf{Model Scalability in Task Performance}.
}
\label{tb:model_scale_task}
\vspace{-2mm}
\begin{tabular*}{\columnwidth}{@{\extracolsep{\fill}}lccc}
\toprule    
 & \multicolumn{3}{c}{\textbf{GQA}} \\
 & Small & Medium & Large \\
\midrule
\multicolumn{4}{@{}l}{\cellcolor{LightRed}\textbf{Qwen3-VL-4B/8B/32B}} \\
Direct & 59.62 & 60.73 & 61.77 \\
CoT & 50.89 & 51.96 & 54.01 \\\midrule
LVLM-CBM & 42.41 & 46.73 & 49.20 \\
RBM & \textbf{46.87} & \textbf{54.89} & \textbf{53.52} \\
\midrule
\multicolumn{4}{@{}l}{\cellcolor{LightBlue}\textbf{Gemma3-4B/12B/27B}} \\
Direct & 45.02 & 44.17 & 52.48 \\
CoT & 39.51 & 24.64 & 31.78  \\\midrule
LVLM-CBM & 33.27 & 12.61 & 26.49 \\
RBM & \textbf{38.84} & \textbf{17.42} & \textbf{39.91} \\
\midrule
\multicolumn{4}{@{}l}{\cellcolor{LightYellow}\textbf{InternVL3.5-4B/8B/38B}} \\
Direct & 60.26 & 61.41 & 64.73 \\
CoT & 51.19 & 54.56 & 57.05 \\\midrule
LVLM-CBM & 43.63 & 45.76 & 53.83 \\
RBM & \textbf{46.38} & \textbf{48.03} & \textbf{55.78} \\
\bottomrule
\end{tabular*}
\vspace{-3mm}
\end{table}

\subsection{Computational Cost}\label{sec:exp_compute_cost}
We compared the inference latency per query in seconds on MMMU-Pro: Direct: 1.17, CoT: 9.72, RBM: 7.89.
RBM's two-stage inference incurs additional latency compared to Direct, but remains competitive with CoT because the second stage omits image tokens. This overhead is a reasonable trade-off for inherent interpretability, aligning with modern test-time scaling trends~\cite{Agarwal_arXiv25_gptoss}.

\section{Limitations}
Despite their advantages, RBMs have two limitations. 
First, the strict rationale bottleneck can discard fine-grained visual information, leading to performance degradation compared to direct inference, particularly in image captioning. 
Second, the two-stage inference and multiple rationale-conditional forward passes introduce additional computational overhead, although RBMs remain faster than standard CoT in our measured setting. 
Future work will focus on preserving informative visual evidence in rationales and improving decoding efficiency to mitigate these trade-offs.

\section{Conclusion}
We presented Relational Bottleneck Models (RBMs), a training-free framework for interpretable inference with off-the-shelf LVLMs. 
By combining Structured Rationale Generation (SRG) and Rationale-Grounded Decoding (RGD), RBMs construct query-conditioned relational rationales and enforce their use in final predictions without directly accessing the input image. 
Experiments across diverse multi-modal benchmarks demonstrate improved rationale interpretability and zero-shot task performance over LVLM-CBM baselines. 
Despite the trade-offs in predictive performance and computational cost, our findings highlight the potential of relational bottlenecks for building interpretable and task-flexible LVLMs.\looseness-1

\clearpage

\section*{LLM Usage Statement}
We used OpenAI's ChatGPT and Codex to assist with manuscript preparation and code development. 
ChatGPT was used for interactive discussions and manuscript refinement, including organization, technical explanations, figure presentation, and wording. 
Codex assisted in writing and modifying implementation and experimental code. 
For the research experiments, we evaluated Qwen3-VL, Gemma 3, and InternVL3.5 as backbone models, used GPT-5-mini for automated rationale evaluation, and used Gemini 3 Flash to generate reference rationales for intervention analysis, as described in Section~\ref{sec:experiments}. 
All AI-assisted content and experimental results were reviewed by the authors, who take full responsibility for the methodology, findings, and conclusions.

{
    \small
    \bibliographystyle{ieeenat_fullname}
    \bibliography{main}
}

\clearpage
\appendix
\onecolumn
\makeatletter
\renewcommand{\maketitlesupplementary}{%
  \begin{center}%
    \bfseries\Large \@title \par
    \vspace{0.5em}
    \large Supplementary Material \par
    \vspace{1.5em}
  \end{center}%
}
\makeatother
\maketitlesupplementary

\section{Details of Benchmark Datasets}\label{append:dataset_details}
\begin{itemize}
    \item \textbf{GQA}~\cite{Hudson_CVPR19_gqa} is a general VQA dataset that contains various questions regarding object/attribute recognition, spatial reasoning, and logical reasoning; we used \texttt{balanced} partitions for testing.
    \item \textbf{TextVQA}~\cite{Singh_CVPR19_textvqa} is a VQA specialized for reading texts in images with OCR tokens given by external models.
    \item \textbf{MME}~\cite{Fu_2023_mme} is a comprehensive binary VQA composed of 14 subtasks to assess both perception and cognition of LVLMs.
    \item \textbf{COCO Caption}~\cite{Lin_2014_COCO} is a representative image captioning dataset. Models require outputting the captions associated with the input images.
    \item \textbf{MMMU}~\cite{Yue_CVPR24_mmmu} and \textbf{MMMU-Pro}~\cite{Yue_ACL25_mmmu-pro} are complex multi-modal reasoning benchmarks that require models to rigorously recognize information from input images and integrate it into the reasoning process. Each instance consists of a question, input images, and multiple-choice candidate answers. Models predict the answer from the multiple-choice.
\end{itemize}

\section{Instruction Prompts for Evaluations}\label{append:prompts}
\subsection{Instruction for CoT Baseline}
We used the following instructions to build the CoT baseline in Section 5:

\begin{figure}[htbp]
\centering
\begin{promptbox}[title={Instruction for CoT Rationale Generation}]
\#\# Instruction

For the provided image and its associated question, please provide detailed reasoning required to output the answer. To output the description, please identify each element related to the question in the image and explain how each contributes to answer the question. Please compose your output with short and clear sentences. 

\#\# Response

[Provide only the necessary and sufficient description here. Be specific and concise.]
\end{promptbox}
\end{figure}

\subsection{Instruction for LVLM-CBMs}
As an inherently interpretable baseline, we built LVLM-CBMs by the following instruction prompts:

\begin{figure}[htbp]
\centering
\begin{promptbox}[title={Instruction for Concept Generation}]
\#\# Instruction

Analyze the image and identify key visual concepts with a relative importance score to answer the question.
Infer the relative importance scores for each rationale based on the contribution to constructing the final answer.
Provide your response in the following JSON Schema:

\{
  "type": "object",
  "properties": \{
    "concepts": \{
      "type": "array",
      "items": \{
        "type": "object",
        "properties": \{
          "concept": \{ "type": "string" \},
          "importance\_score": \{
            "type": "number",
            "minimum": 0,
            "maximum": 1
          \}
        \},
        "required": ["concept", "importance\_score"]
      \}
    \}
  \}
\}

\#\# Response

[Identify up to $K$ most relevant concepts here.]
\end{promptbox}
\end{figure}

The main difference from RBMs is that this instructs model to generate only visual concepts for answering the given question.
We also generated importance scores for each concept to ablate the concept format in RBMs in Section 5.4.

\subsection{Instruction for Rationale Accuracy}
To automatically evaluate the Rationale Accuracy of generated rationales, we gave the judge model (GPT-5-mini) the following instructions:

\begin{figure}[htbp]
\centering
\begin{promptbox}[title={LLM-as-a-Judge Instruction for Rationale Accuracy Evaluation}]
You are an expert evaluator for visual question answering (VQA) rationales.
Given a question, a predicted answer, and a rationale, evaluate the rationale on
three criteria and assign an integer score from 1 to 5 for each.

Criteria:

1. Faithfulness (1–5)\\
\ptab Does the rationale make accurate and consistent claims about the visual scene?\\
\ptab 5 – All claims are accurate and fully grounded in the visual scene.\\
\ptab 4 – Mostly accurate with only minor imprecisions that do not mislead.\\
\ptab 3 – Partially accurate; some claims are correct but notable inaccuracies exist.\\
\ptab 2 – Mostly inaccurate or contains significant hallucinations.\\
\ptab 1 – Completely hallucinated, contradictory, or implausible descriptions.\\

2. Query-Relevance (1–5)\\
\ptab Does the rationale provide information that enables deriving the predicted answer?\\
\ptab 5 – Directly and completely supports the answer; no irrelevant content.\\
\ptab 4 – Mostly relevant with minor tangential content.\\
\ptab 3 – Partially relevant; provides some useful evidence but also irrelevant material.\\
\ptab 2 – Marginally relevant; little connection to the answer.\\
\ptab 1 – Entirely unrelated or actively misleading with respect to the answer.\\

3. Coverage (1–5)\\
\ptab Does the rationale comprehensively identify the key visual evidence needed?\\
\ptab 5 – All critical visual evidence is identified; nothing important is missing.\\
\ptab 4 – Most key evidence is covered; only minor details are omitted.\\
\ptab 3 – Core evidence is partially covered; some important details are missing.\\
\ptab 2 – Coverage is sparse; several critical pieces of evidence are absent.\\
\ptab 1 – Nearly no relevant visual evidence is identified; rationale is too vague.\\

Respond ONLY with a JSON object in the following format (no markdown fences):\\
\{\\
\ptab "faithfulness": <1–5>,\\
\ptab "faithfulness\_explanation": "<brief explanation>",\\
\ptab "query\_relevance": <1–5>,\\
\ptab "query\_relevance\_explanation": "<brief explanation>",\\
\ptab "coverage": <1–5>,\\
\ptab "coverage\_explanation": "<brief explanation>"\\
\}\\

\#\# Question\\
\{question\}\\

\#\# Predicted Answer\\
\{answer\}\\

\#\# Rationale\\
\{rationale\}\\

Evaluate the rationale on the three criteria described in the above.
\end{promptbox}
\end{figure}

All of the perspectives of faithfulness, query relevance, and coverage are quantified in 1-5 scores and higher is better.

\subsection{Instruction for Rationale Preference}
We automatically evaluated the Rationale Preference scores by the following prompts:

\begin{figure}[htbp]
\centering
\begin{promptbox}[title={LLM-as-a-Judge Instruction for Rationale Preference Evaluation}]
You are an expert evaluator assessing the quality of rationales used to answer visual question answering (VQA) questions.
Your task is to determine which of two given rationales provides more interpretable information for correctly deriving the answer to the question.
A rationale is considered more interpretable if it:\\
- Identifies the key visual features that are directly relevant to answering the question.\\
- Contains accurate and specific evidence that logically leads to the correct answer.\\
- Avoids irrelevant or misleading information.\\

Respond ONLY with a JSON object in the following format (no markdown fences):\\
\{\\
\ptab "preference": "<A|B|tie>",\\
\ptab "explanation": "<brief explanation of your judgment>"\\
\}
\\
\\
\#\# Question\\
\{question\}\\

\#\# Rationale A\\
\{rationale\_a\}\\

\#\# Rationale B\\
\{rationale\_b\}\\

Which rationale (A or B) provides more beneficial information for answering the question? \\
If both are equally useful, respond with "tie".
\end{promptbox}
\end{figure}

We randomly swap the order of the rationales for each evaluation iteration in order to avoid the positional bias in LLM-as-a-Judge by following~\cite{Zheng_NeurIPS23_judging-llm-as-a-judge}.

\section{Validity Evaluation on LLM-as-a-Judge}
To validate the trustworthiness of LLM-as-a-Judge for rational accuracy/preference evaluations, we conducted human evaluations of our LLM-as-a-Judge setup on the GQA validation set (100 samples) by two human evaluators.
Specifically, we achieved a Spearman's rank correlation coefficient of 0.70 for the absolute evaluation of rationale accuracy and an agreement rate of 96\% for the relative evaluation.
These results suggest the validity of our evaluations.

\section{Additional Experiments}\label{append:experiments}

\subsection{Comparison to Classification-based CBMs}
Direct comparison of RBMs to most existing classification-based CBMs, such as LF-CBMs~\cite{Oikarinen_ICLR23_label-free_CBMs}, is difficult because the CBMs usually require task-specific training on the target datasets and different architectures, e.g., CLIP~\cite{Radford_ICML21_CLIP}.
Nevertheless, as a reference, we compared RBMs with Z-CBMs~\cite{Yamaguchi_2025_zcbm}, which can perform without any additional training.
We used Z-CBM built on CLIP-B/32 and compared it with RBMs on the CUB dataset.
RBMs outperformed Z-CBMs in accuracy (\textbf{68.02} vs. 51.75), highlighting their practicality.

\end{document}